\documentclass[10pt,twocolumn,letterpaper]{article}

 \usepackage[pagenumbers]{iccv} 

\usepackage{multirow}
\usepackage{pifont}

\definecolor{iccvblue}{rgb}{0.21,0.49,0.74}
\usepackage[pagebackref,breaklinks,colorlinks,allcolors=iccvblue]{hyperref}

\def\paperID{4978} 
\def\confName{ICCV}
\def\confYear{2025}

\usepackage{colortbl}
\usepackage{xcolor}

\title{End-to-End Multi-Person Pose Estimation with Pose-Aware Video Transformer}

\author{
Yonghui Yu\thanks{Equal contribution.}, 
Jiahang Cai\footnotemark[1], 
Xun Wang,
Wenwu Yang\thanks{Corresponding author (wwyang@zjgsu.edu.cn).}\\[5pt]
Zhejiang Gongshang University, China\\
}

\begin{document}
\maketitle
\begin{abstract}

Existing multi-person video pose estimation methods typically adopt a two-stage pipeline: detecting individuals in each frame, followed by temporal modeling for single-person pose estimation. This design relies on heuristic operations such as detection, RoI cropping, and non-maximum suppression (NMS), limiting both accuracy and efficiency. 
In this paper, we present a fully end-to-end framework for multi-person 2D pose estimation in videos, effectively eliminating heuristic operations. A key challenge is to associate individuals across frames under complex and overlapping temporal trajectories. 
To address this, we introduce a novel Pose-Aware Video transformEr Network (PAVE-Net), which features a spatial encoder to model intra-frame relations and a spatiotemporal pose decoder to capture global dependencies across frames. To achieve accurate temporal association, we propose a pose-aware attention mechanism that enables each pose query to selectively aggregate features corresponding to the same individual across consecutive frames.
Additionally, we explicitly model spatiotemporal dependencies among pose keypoints to improve accuracy.
Notably, our approach is the first end-to-end method for multi-frame 2D human pose estimation.
Extensive experiments show that PAVE-Net substantially outperforms prior image-based end-to-end methods, achieving a \textbf{6.0} mAP improvement on PoseTrack2017, and delivers accuracy competitive with state-of-the-art two-stage video-based approaches, while offering significant gains in efficiency. 
Project page: https://github.com/zgspose/PAVENet

\end{abstract}    
\section{Introduction}
\label{sec:intro}

Multi-person 2D pose estimation (MPPE) aims to detect and localize anatomical keypoints of all individuals in images or videos, and is fundamental to many applications such as human-computer interaction~\cite{HOI-CVPR2020}, behavior analysis~\cite{ActionRecog-PAMI2023}, and motion capture~\cite{MOCap-CVPR2020}.
Although MPPE is widely used in videos, most existing methods process frames independently as static images~\cite{TokenPose_CVPR2023,CID_cvpr2022,Vitpose_NIPS2022,HRNet_CVPR2019,SimplePose_ECCV2018,RLE_ICCV2021,PEDR-CVPR2022,PoseLLM-CVPR2024,Sapiens-ECCV2024}.
Recent works~\cite{PoseWarper_NIPS2019,DCPose_CVPR2021,OTPose_SMC2022,TDMI_CVPR2023,DSTA-CVPR2024} incorporate temporal information to better handle occlusion, motion blur, and defocus, 
highlighting the importance of video dynamics.

\begin{figure}
    \centering
    \includegraphics[width=1.0\linewidth]{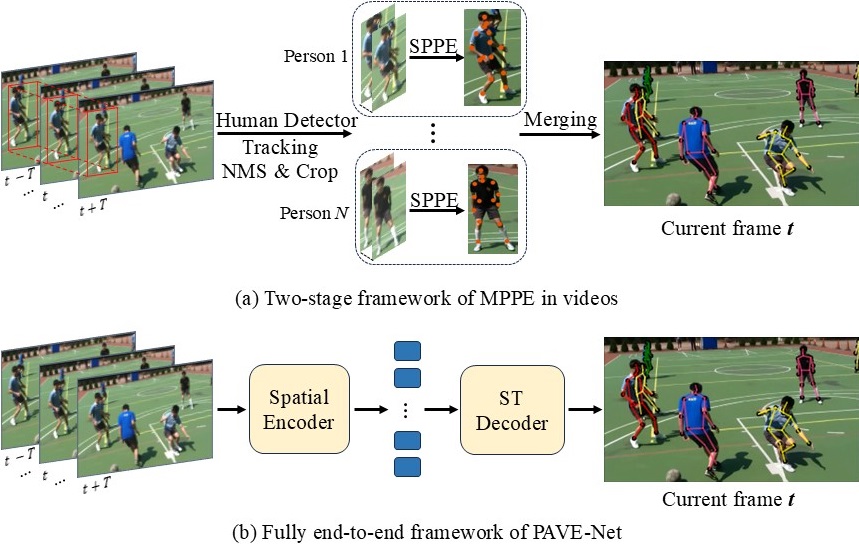}
    \caption{ Comparison of two-stage and end-to-end frameworks for video-based MMPE. (a) To predict 2D poses in the current frame, existing methods~\cite{PoseWarper_NIPS2019, DCPose_CVPR2021, OTPose_SMC2022, TDMI_CVPR2023, DiffPose-ICCV2023, DSTA-CVPR2024} first crop regions from consecutive frames for each human instance and then input them into a temporal model to perform single-person pose estimation (SPPE). (b) PAVE-Net achieves end-to-end video-based MMPE with a spatial encoder and spatiotemporal decoder. }
    \label{fig:demo}
    \vspace{-7mm}
\end{figure}

Existing video-based methods all follow a detection-based pipeline, as shown in Fig.~\ref{fig:demo}(a), where human instances are first detected and their regions cropped from consecutive frames before being fed into a CNN-based~\cite{PoseWarper_NIPS2019,DCPose_CVPR2021,TDMI_CVPR2023,DiffPose-ICCV2023} or Transformer-based~\cite{OTPose_SMC2022,DSTA-CVPR2024} temporal model for single-person pose estimation.
This framework has several drawbacks: (1) it fails to capture spatial relations among human instances, (2) its accuracy heavily depends on human detector performance~\cite{CID_cvpr2022,PEDR-CVPR2022}, leading to suboptimal results in crowded scenarios, and (3) it is computationally expensive due to the separate detector, with runtime increasing as the number of people grows.
Thus, these methods split the task into two stages, preventing full end-to-end optimization.
To overcome the limitations of two-stage pipelines, several fully end-to-end frameworks have recently been proposed for image-based MPPE~\cite{PEDR-CVPR2022,EDPose-ICLR2023,GroupPose-ICCV2023}. These methods typically adopt an encoder-decoder transformer architecture: the encoder captures local dependencies among image feature tokens, while the decoder uses pose queries to directly infer full-body poses.
A straightforward extension to video-based MPPE, as explored in~\cite{PSVT_cvpr2023} for 3D pose estimation, incorporates temporal information via a spatiotemporal encoder, modeling global dependencies among feature tokens across frames. However, this extension dramatically increases computational complexity, as transformer attention scales quadratically with token count; for example, processing 5 frames increases the computational load by 25×, leading to prohibitive memory and compute costs.
Furthermore, as illustrated in Fig.~\ref{fig:demo}, individuals in multi-person video scenes follow independent, potentially overlapping temporal trajectories, posing a significant challenge: accurately associating identities over time is critical to prevent feature mixing and to enable effective temporal aggregation.

To effectively address these challenges, we propose PAVE-Net, a fully end-to-end framework specifically designed for multi-person 2D pose estimation in videos.
PAVE-Net integrates spatial-temporal dependency modeling at both the instance level (individuals) and the joint level (fine-grained body keypoints).
As shown in Fig.~\ref{fig:demo}(b), PAVE-Net takes as input the current frame along with several adjacent frames and outputs 2D poses for all individuals in the current frame.
Specifically, it first encodes local dependencies among visual feature tokens from each frame. Next, a spatiotemporal pose decoder captures global dependencies between pose queries and tokens across frames. To ensure each pose query accurately aggregates features corresponding to the same individual over time, we introduce a novel pose-aware attention mechanism that predicts initial pose estimates and uses them to guide query-to-feature matching across frames.
Finally, a spatiotemporal joint decoder explicitly analyzes dependencies among keypoints in each pose, further refining the multi-person pose estimates.

To the best of our knowledge, this is the first fully end-to-end framework for multi-person 2D pose estimation in videos. We extensively evaluate our method on three widely used video-based MPPE benchmarks: PoseTrack2017~\cite{PoseTrack2017_CVPR2017}, PoseTrack2018~\cite{PoseTrack2018_CVPR2018}, and PoseTrack21~\cite{PoseTrack21_CVPR2022}.
Experimental results show that our framework achieves substantial improvements, outperforming prior end-to-end image-based methods by a notable margin (\textit{e.g.}, \textbf{6.0} mAP gain), while delivering accuracy comparable to state-of-the-art two-stage methods.
Moreover, unlike two-stage methods, our method eliminates the human detection step, removing heuristic operations such as RoI cropping and NMS, offering significant efficiency gains and full end-to-end differentiability. It should be noted that our work does not address temporal pose tracking, but rather presents an end-to-end framework with superior performance for 2D pose estimation in videos.


Our main contributions can be summarized as follows: 
\begin{itemize}
    \item We propose \textbf{PAVE-Net}, a novel end-to-end framework for multi-person video pose estimation that efficiently and flexibly models spatiotemporal relationships among both human instances and fine-grained body joints.
    \item  Our method is the first end-to-end approach for multi-frame, multi-person 2D pose estimation. Unlike existing two-stage methods, it directly predicts instance-aware full-body poses, eliminating the need for human detection, RoI cropping, and NMS. 
    \item Extensive experiments demonstrate that our method significantly outperforms prior end-to-end image-based MPPE approaches, while achieving accuracy comparable to state-of-the-art two-stage video-based methods and offering substantial efficiency gains. 
\end{itemize}

\section{Related Work}
\label{sec:relate_work}

Multi-person pose estimation (MPPE) is a challenging task as it requires not only accurate recognition of each individual but also the ability to handle varying poses, occlusions, and background interference in images or videos. 
Existing methods can be broadly categorized into two types: non-end-to-end methods and end-to-end methods.

\textbf{Non-end-to-end MPPE methods} can be further classified into two main types: two-stage and one-stage approaches. Among the two-stage methods, there are two subcategories: top-down and bottom-up methods. In the top-down approach~\cite{Vitpose_NIPS2022,HRNet_CVPR2019,SimplePose_ECCV2018,RLE_ICCV2021,Poseur_ECCV2022,RMPE_ICCV2017,CPN-CVPR2018}, a human detector first generates bounding boxes for each person, which are then cropped and passed to a single-person pose estimation model. In contrast, the bottom-up method~\cite{OpenPose-PAMI2021,PersonLab-ECCV2018,PifPaf-CVPR2019,HigherHRNet-CVPR2020} predicts keypoints for all individuals simultaneously and uses a grouping algorithm to associate them with their respective human instances. On the other hand, one-stage methods~\cite{CID_cvpr2022,SPM-ICCV2019,InsPose-MM2021,RTMO-CVPR2024} directly predict the keypoints for all human instances in the image. Both two-stage and one-stage methods rely on manually designed preprocessing or postprocessing steps, such as NMS or grouping algorithms, which prevent them from achieving end-to-end optimization. Recently, complementary information from adjacent frames has been leveraged to capture temporal dependencies in human poses, thereby improving the performance of MMPE in video sequences. However, all these video-based methods~\cite{PoseWarper_NIPS2019, DCPose_CVPR2021, OTPose_SMC2022,FAMIPose_CVPR2022,TDMI_CVPR2023, DiffPose-ICCV2023, DSTA-CVPR2024} follow a top-down framework, making their performance highly dependent on the human detector and significantly affecting inference speed as the number of detected human instances increases. 

\textbf{End-to-end MPPE methods} are mostly built upon the end-to-end object detection framework DETR~\cite{DETR-Arxiv2020} and its variants. These methods eliminate the need for hand-crafted operations such as RoI cropping, NMS, and grouping. They typically decompose the MPPE task into two main subtasks. For example, \cite{PEDR-CVPR2022} formulates it as a hierarchical set prediction problem, first using a pose decoder to predict a set of poses and identify human instances, then refining the keypoints of each pose with a keypoint decoder. \cite{QueryPose_NIPS2022,EDPose-ICLR2023} follow this end-to-end paradigm but further incorporate an additional human detection task to provide a better initialization for subsequent keypoint detection. In~\cite{GroupPose-ICCV2023}, the pose decoder and keypoint decoder are integrated into a single unified decoder, enabling interaction between human instances and their keypoints, as well as interaction among the same keypoints across different human instances. However, these end-to-end methods are designed specifically for static images, and extending them to videos is non-trivial due to the added complexity of the temporal dimension. When applied directly to video sequences, these image-based  methods often produce suboptimal results due to their inability to capture temporal dependencies between frames, making them ineffective in handling challenges like motion blur, defocus, and occlusions common in dynamic scenes.

In this work, we introduce the first end-to-end framework for multi-person pose estimation in videos, which competes favorably with state-of-the-art two-stage top-down video-based methods in terms of both accuracy and efficiency.

\section{Method}

Given a current video frame \({F}(t)\) at time \(t\) containing multiple individuals, our goal is to estimate the locations of pose joints for each person by leveraging temporal dynamics from a sequence of consecutive frames \(\langle {F}({t-T}), \cdots, {F}(t), \cdots, {F}({t+T}) \rangle\), where \(T\) denotes a predefined temporal span. 
Our method adopts the encoder-decoder transformer architecture commonly used in end-to-end image-based MPPE~\cite{PEDR-CVPR2022,EDPose-ICLR2023,GroupPose-ICCV2023}: feature tokens are extracted from video frames by the encoder, followed by a pose decoder that learns multiple pose queries to directly predict full-body poses. 
A joint decoder further refines the predictions at the joint level. 
In contrast to prior end-to-end MPPE models for static images~\cite{PEDR-CVPR2022,EDPose-ICLR2023,GroupPose-ICCV2023}, our method introduces a video-based end-to-end framework that effectively and efficiently exploits temporal information.

\subsection{Video Transformer Baseline}
\label{sec:baseline}
\begin{figure*}
    \centering
    \includegraphics[width=1.0\linewidth]{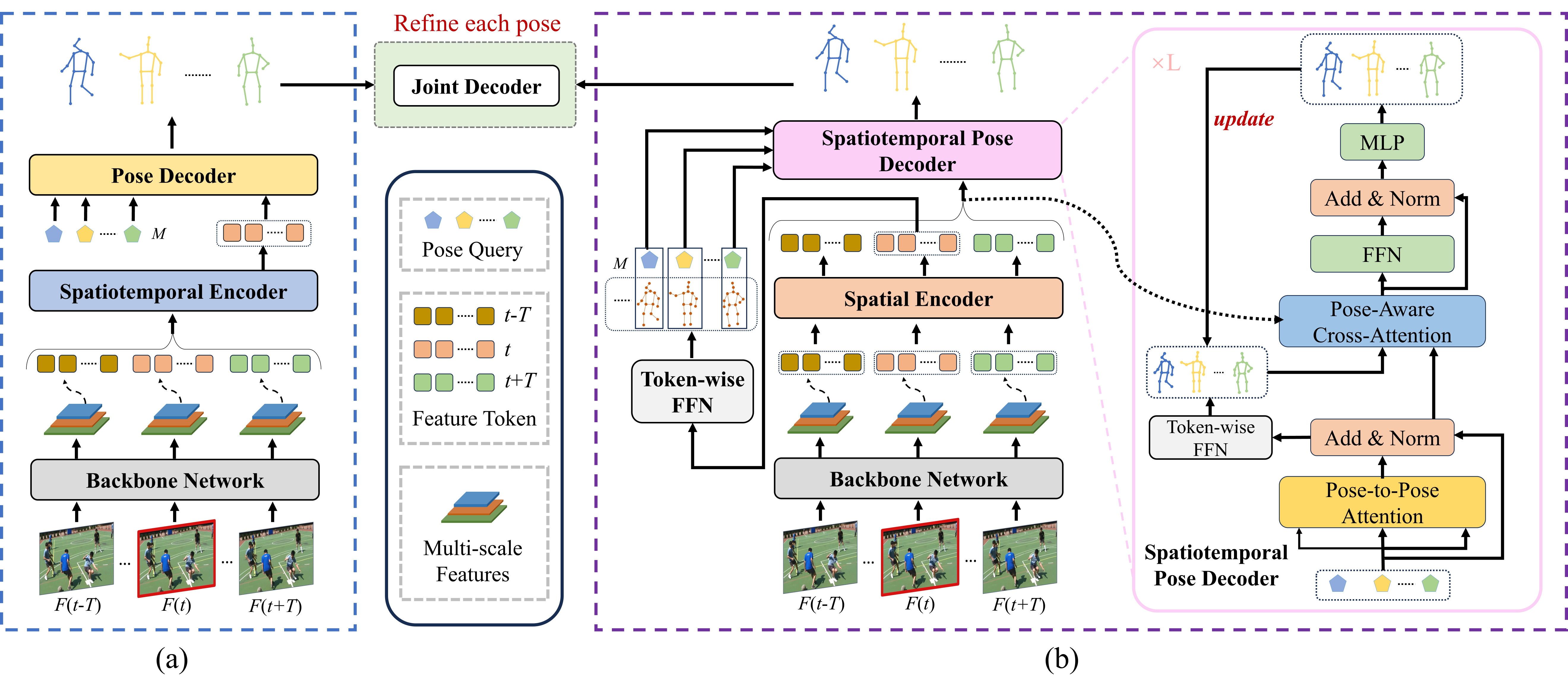}
    \vspace{-7mm}
    \caption{(a) Video transformer baseline. (b) Pose-aware video transformer (PAVE-Net) architecture. The goal is to detect all human poses in the current frame \( F(t) \) by leveraging temporal dynamics from a sequence of consecutive frames \(\langle F(t-T), \dots, F(t), \dots, F(t+T) \rangle\). PAVE-Net employs a backbone network to extract multi-scale features from each frame, which are transformed into feature tokens. A Spatial Encoder (SE) processes each frame independently to capture local dependencies within its tokens. The Spatiotemporal Pose Decoder (STPD) then models global dependencies between pose queries and feature tokens across all frames, using the top \( M \) highest-confidence poses regressed from the feature tokens of the current frame \( t \) as references. This enables accurate prediction of 2D poses for frame \( t \), which are further refined by a joint decoder.
 }
    \label{fig:pipeline}
    \vspace{-5mm}
\end{figure*}

As a baseline application of the encoder-decoder transformer architecture in end-to-end video-based MPPE, as adopted in~\cite{PSVT_cvpr2023} for 3D pose estimation, we employ a spatiotemporal encoder to capture global dependencies among visual feature tokens extracted from all input video frames, as illustrated in Fig.~\ref{fig:pipeline}. 
For the input frame sequence \( X \in \mathbb{R}^{f \times H \times W \times 3} \), \( f \) denotes the number of frames, \( H \) and \( W \) the height and width of each frame, and 3 the number of color channels. 
Each frame is represented as \( F({t'}) \in \mathbb{R}^{H \times W \times 3} \), where \( {t'} \in [t-T, t+T] \). 
For each \( F({t'}) \), multi-scale features \( Z({t'}) \in \mathbb{R}^{\frac{H}{s} \times \frac{W}{s} \times C_s} \) are extracted by a backbone network (\textit{e.g.}, ResNet~\cite{ResNet-CVPR2016}), where \( s \) is the scale factor and \( C_s \) the feature dimensionality at that scale. 
\( Z({t'}) \) is then converted into tokens \(\tau({t'}) \in \mathbb{R}^{N \times D} \) via patch embedding with a patch size of \( 1 \times 1 \), where \( N \) is the number of tokens and \( D \) the embedding dimension. 
After concatenating tokens from all frames, the sequence \( X \in \mathbb{R}^{f \times H \times W \times 3} \) is transformed into \( Y \in \mathbb{R}^{f \times N \times D} \), which is fed into the spatiotemporal transformer encoder.


\textbf{Spatiotemporal Transformer Encoder} captures global dependencies among feature tokens across spatial and temporal dimensions,
\begin{equation}
\label{eqn:STE}
    \Tilde{\tau}(t) = \text{STE}( Y ),
\end{equation}
where $\text{STE}(\cdot)$ denotes the spatiotemporal encoder module, and $\Tilde{\tau}(t) \in \mathbb{R}^{N \times D}$ represents the updated feature tokens for the current video frame $t$, encoding both spatial and temporal information learned by the STE. 
In our implementation, we adopt deformable attention~\cite{DETR-Arxiv2020} for an efficient realization of the STE module, rather than standard multi-head self-attention~\cite{Attention_NIPS2017}. 
We stack six identical deformable attention layers, each consisting of a multi-scale deformable attention module and a simple token-wise feed-forward network. 
Feature tokens pass through these layers sequentially, with each producing an updated version that serves as input for the next. 
Additionally, each initial feature token is augmented with a learnable position embedding and a feature scale-level embedding, whose sum forms the encoder input.

\textbf{Pose Decoder} computes cross-attention between \( M \) learnable pose query tokens \( Q^p \in \mathbb{R}^{M \times D} \) and the feature tokens of the current video frame \( \tilde{\tau}(t) \in \mathbb{R}^{N \times D} \), which encode both spatial and temporal information,
\begin{equation}
\label{eqn:PD}
    \Tilde{Q}^p = \text{PD}( Q^p, \tilde{\tau}(t) ),
\end{equation}
where $\text{PD}(\cdot)$ denotes the pose decoder module, and \( \tilde{Q}^p \) are the updated query tokens, each extracting features from \( \tilde{\tau}(t) \). 
Following the STE module, we adopt deformable attention to build the PD module for efficiency. 
As in~\cite{PEDR-CVPR2022}, we stack three identical deformable cross-attention layers. 
In each layer, query tokens first interact through a self-attention module (\textit{i.e.}, pose-to-pose attention), followed by deformable cross-attention to extract features from \( \tilde{\tau}(t) \) (\textit{i.e.}, feature-to-pose attention).
To predict full-body poses at frame \( t \), the decoder output \( \tilde{Q}^p \) is passed through two token-wise feed-forward networks: one predicts \( M \) full-body poses and the other predicts confidence scores for each pose,
\begin{equation}
\label{eqn:head}
    \left\{
\begin{aligned}
 \tilde{Q}^p \xrightarrow[\text{feed-forward network}]{\text{token-wise fully connected}} \{ P_i(t)  \}_{i=1}^M \\
 \tilde{Q}^p \xrightarrow[\text{feed-forward network}]{\text{token-wise fully connected}} \{ c_i(t)  \}_{i=1}^M
\end{aligned}
\right.,
\end{equation}
where \( P_i(t) \in \mathbb{R}^{J \times 2} \) denotes the joint coordinates of the \( i \)-th 2D pose, with \( J \) representing the number of joints (\textit{e.g.}, \( J=15 \) for the PoseTrack datasets~\cite{PoseTrack2017_CVPR2017}), and \( c_i(t) \) is the confidence score for the \( i \)-th pose. 
A joint decoder, discussed in the next section, is further used to refine predicted poses at the joint level.

\subsection{Pose-Aware Video Transformer}

We observe that the video transformer baseline primarily relies on the spatiotemporal encoder \( \text{STE}(\cdot) \) to capture global dependencies across frames and leverage temporal information. 
However, compared to the encoder used in end-to-end image-based MPPE methods~\cite{PEDR-CVPR2022,EDPose-ICLR2023,GroupPose-ICCV2023}, which only captures local dependencies within a single image, the computational complexity of the spatiotemporal encoder increases substantially. 
For example, when \textit{T} = 2  (\textit{i.e.}, using 5 frames), its computational demand becomes 25 times higher than that of a single-frame encoder. 
In addition, multi-person video scenarios introduce a key technical challenge: temporal features from different individuals can easily become entangled without explicit cross-frame associations. 
Therefore, accurately associating identities across frames is critical for effectively aggregating temporal features corresponding to each individual.

To effectively and efficiently exploit temporal information in videos, we propose a novel Pose-Aware Video transformEr network, called PAVE-Net. 
As illustrated in Fig.~\ref{fig:pipeline}, PAVE-Net consists of three main modules: the spatial encoder (SE), the spatiotemporal pose decoder (STPD), and the spatiotemporal joint decoder (STJD).

\textbf{Spatial Encoder.} 
Unlike the \( \text{STE}(\cdot) \) module used in the baseline, which captures global dependencies among feature tokens across multiple frames, the spatial encoder (SE) processes each frame independently to capture local dependencies within its feature tokens,
\begin{equation}
\label{eqn:SE}
    \hat{\tau}(t') = \text{SE}( \tau(t') ), \quad t' \in [t-T, t+T],
\end{equation}
where \( \text{SE}(\cdot) \) denotes the spatial encoder module, and \( \hat{\tau}(t') \in \mathbb{R}^{N \times D} \) represents the updated feature tokens for frame \( t' \), encoding spatial information learned by SE. 
Since SE captures only local dependencies within each frame and operates independently across frames, its output can be reused for different frame sequences. 
For example, \( \hat{\tau}(t) \) can be reused for pose estimation from frame \( t-T \) to \( t+T \). 
As a result, the computational complexity of SE is equivalent to that of a single-frame encoder. 
In our implementation, SE adopts the same architecture as the spatiotemporal encoder (STE) in Eq.~\ref{eqn:STE}, consisting of multi-scale deformable attention and feed-forward network (FFN) blocks.

\textbf{Spatiotemporal Pose Decoder with Pose-Aware Attention.} 
Since each set of feature tokens \( \hat{\tau}(t') \), where \( t' \in [t-T, t+T] \), encodes only local dependencies within frame~\( t' \), the spatiotemporal pose decoder (STPD) must compute cross-attention between the \( M \) learnable pose query tokens \( Q^p \in \mathbb{R}^{M \times D} \) and the feature tokens from all input frames to aggregate temporal features,
\begin{equation}
    \hat{Q}^p = \text{STPD}( Q^p, \{\hat{\tau}(t')\}_{t'=t-T}^{t+T} ),
\end{equation}
where \( \text{STPD}(\cdot) \) denotes the spatiotemporal pose decoder, and \( \hat{Q}^p \) represents the updated query tokens, each aggregating features corresponding to the same individual across frames. 
STPD adopts a similar architecture to the pose decoder (PD) in Eq.~\ref{eqn:PD}, consisting of three stacked layers, each with self-attention among query tokens and deformable cross-attention between query tokens and multi-scale feature tokens. 
To ensure that each pose query token consistently aggregates features from the same individual across frames, we introduce a pose-aware attention mechanism, which guides each query to attend only to feature tokens associated with the same person throughout the temporal window.

We first predict a set of initial poses from the feature tokens \( \hat{\tau}(t) \) of the current frame \( t \), where each feature token is used to regress a full-body pose and its confidence score via two token-wise regression heads, similar to Eq.~\ref{eqn:head}. 
The \( M \) poses with the highest confidence scores are then selected as reference poses, with each pose query token assigned a corresponding reference. 
Note that, unlike~\cite{PEDR-CVPR2022}, our method does not assign query tokens to a fixed set of randomly initialized reference points.

For each pose query token \( \textbf{q}^p_i \in \mathbb{R}^D \), \( i=1,2,\dots,M \), we denote \( P^0_i(t) \in \mathbb{R}^{J \times 2} \) as its initial reference pose. 
Since an individual’s pose remains similar across adjacent frames, we reuse \( P^0_i(t) \) as the reference for that individual across all input frames, enabling us to locate corresponding features in each frame. 
A token-wise FFN regresses relative offsets for each input frame with respect to the reference pose,
\begin{equation}
\label{eqn:offset}
\textbf{q}^p_i \xrightarrow[\text{FFN}]{\text{token-wise}} \{ \Delta P_i(t')  \}_{t'=t-T}^{t+T},
\end{equation}
where \( \Delta P_i(t') \) is the relative offset at frame \( t' \). 
Thus, for each pose query token \( \textbf{q}^p_i \), the target positions at frame \( t' \) are given by \( P^0_i(t') + \Delta P_i(t') \), where \( P^0_i(t') = P^0_i(t) \) for all \( t' \). 
The query token then extracts relevant features from \( \{ \hat{\tau}(t') \}_{t'=t-T}^{t+T} \) at these positions, performing cross-attention to aggregate features corresponding to the same individual across frames.
Inspired by~\cite{DETR-Arxiv2020, PEDR-CVPR2022}, we progressively refine reference poses across decoder layers: \( P^l_i(t') = P^{l-1}_i(t') + \Delta P^l_i(t') \), where \( \Delta P^l_i(t') \) is predicted by feeding the updated query token into a token-wise FFN, and \( P^{l-1}_i(t') \) serves as the reference in the \( l \)-th decoder layer. 
The final predicted 2D poses for frame \( t \) are \( \{ P_i(t) \}_{i=1}^M \), where each \( P_i(t) = P^3_i(t) \).

\textbf{Spatiotemporal Joint Decoder.} 
The joint decoder captures kinematic dependencies between articulated joints, further refining each predicted pose at the joint level. 
For each pose \( P_i(t) \) predicted by the spatiotemporal pose decoder, the joint decoder uses its joint locations as initial reference points and refines them via cross-attention between a shared set of \( J \) learnable joint query tokens \( Q^o \in \mathbb{R}^{J \times D} \) and the feature tokens from all input frames,
\begin{equation}
    \hat{Q}^o = \text{STJD}( Q^o, \{\hat{\tau}(t')\}_{t'=t-T}^{t+T} \vert P_i(t) ),
\end{equation}
where \( \text{STJD}(\cdot) \) denotes the spatiotemporal joint decoder, and \( \hat{Q}^o \) are the updated joint query tokens, each aggregating features for the corresponding joint across frames. 
STJD has three layers and follows the architecture of STPD: each layer applies self-attention among joint queries (joint-to-joint attention) followed by deformable cross-attention between joint queries and multi-scale feature tokens (feature-to-joint attention). 
As in STPD, reference points are progressively refined layer by layer, with relative offsets regressed by a token-wise FFN from the updated joint query tokens.

Note that it is straightforward to extend the spatiotemporal joint decoder to the previously described video transformer baseline by replacing the spatial feature tokens of all input frames \( \{ \hat{\tau}(t') \}_{t'=t-T}^{t+T} \) learned by the SE module in Eq.~\ref{eqn:SE} with the spatiotemporal feature tokens of the current frame \( \Tilde{\tau}(t) \) learned by the STE module in Eq.~\ref{eqn:STE}.

\subsection{Loss Function}

During training, the entire model is optimized end-to-end by minimizing the discrepancy between the ground-truth poses and all predicted poses at different stages, including the initial pose and its successive refinements for the current frame.
Similar to~\cite{PEDR-CVPR2022}, we adopt a set-based Hungarian loss to enforce a one-to-one assignment between predictions and ground-truth poses. 
Following~\cite{DSTA-CVPR2024, RLE_ICCV2021}, we replace conventional regression losses (\( l_1 \) or \( l_2 \)) with the residual log-likelihood estimation loss (\( \mathcal{L}_{rle} \)) for pose regression in the pose decoder and joint regression in the joint decoder. 
Additionally, we use the same classification loss (\( \mathcal{L}_{cls} \)) as in~\cite{DETR-Arxiv2020} for regressing pose confidence scores. 
The total loss \( \mathcal{L} \) is defined as
\begin{equation}
  \mathcal{L} =  \lambda_{cls} \mathcal{L}_{cls} + \lambda_{rle} \mathcal{L}_{rle},
\end{equation}
where \( \lambda_{cls} \) and \( \lambda_{rle} \) are weighting factors.

\section{Experiments}
\subsection{Experimental Settings}

We evaluated our model on three widely used video benchmark datasets: PoseTrack2017~\cite{PoseTrack2017_CVPR2017}, PoseTrack2018~\cite{PoseTrack2018_CVPR2018}, and PoseTrack21~\cite{PoseTrack21_CVPR2022}. 
Each dataset contains dynamic video sequences with complex scenes, including significant occlusions and rapid motion in crowded environments. 
To assess performance, we used the Average Precision (AP) metric~\cite{HRNet_CVPR2019,PoseWarper_NIPS2019,DCPose_CVPR2021,RLE_ICCV2021}, where AP is computed for each keypoint and the mean Average Precision (mAP) is obtained by averaging AP over all keypoints. Each result is obtained through 2–4 runs.
Our model was implemented in PyTorch, with the backbone network pre-trained on the COCO dataset. 
The temporal span \( T \) was set to 1, \textit{i.e.}, one preceding frame and one subsequent frame, totaling two auxiliary frames. 
For additional implementation details, please refer to Appendix  B of the supplementary material.


\subsection{Comparison with State-of-the-art Methods}
We begin with a comprehensive performance comparison against state-of-the-art methods on the PoseTrack2017 dataset, and subsequently extend our evaluation to the PoseTrack2018 and PoseTrack21 datasets.

\subsubsection{Results on the PoseTrack2017 Dataset\\}
\label{sec:main_result}


\textbf{Comparison with Image-based End-to-End Methods.} 
To thoroughly evaluate the effectiveness of our proposed end-to-end method for video input, we first compare it against state-of-the-art image-based end-to-end methods, specifically PETR~\cite{PEDR-CVPR2022} and GroupPose~\cite{GroupPose-ICCV2023}. 
To ensure a comprehensive evaluation, we employ three backbone networks: ResNet-50, HRNet-W48, and Swin-L, applying each approach using identical pre-trained models for these backbones. 
Additionally, we replace the conventional regression losses (\( l_1 \) or \( l_2 \)) with the residual log-likelihood estimation loss in our re-implementations of PETR and GroupPose to ensure consistency and fairness.

\textit{Quantitative results:} 
As shown in Table~\ref{tab:sota_compare}, our video-based method consistently achieves substantial performance gains across all backbones. 
For example, it surpasses PETR~\cite{PEDR-CVPR2022} by \textbf{6.0} mAP using ResNet-50 and by \textbf{4.7} mAP using HRNet-W48. 
These results highlight the importance of leveraging temporal cues from adjacent frames, which image-based methods inherently lack.

\begin{figure}
    \centering
    \includegraphics[width=1\linewidth]{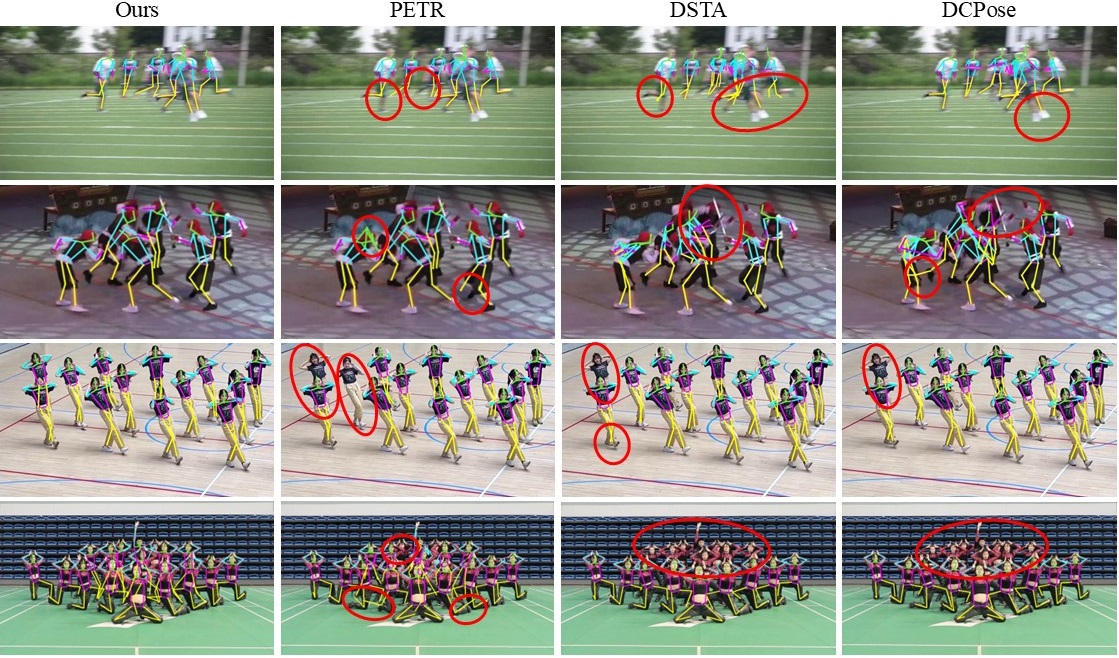}
    \caption{Qualitative comparison of our PAVE-Net, PETR~\cite{PEDR-CVPR2022}, DSTA~\cite{DSTA-CVPR2024}, and DCPose~\cite{DCPose_CVPR2021}, highlighting challenges such as occlusions, motion blur, and crowded scenarios. The top two rows are from the PoseTrack dataset, while the bottom two rows are from in-the-wild videos.
    Inaccurate predictions are marked with red solid circles. Better viewed with zoom.}
    \label{fig:qualitative_comparison}
    \vspace{-2mm}
\end{figure}

\textit{Qualitative results:} 
By effectively leveraging temporal dependencies across consecutive frames, our video-based end-to-end framework demonstrates improved robustness in challenging scenarios such as occlusions and motion blur, which are common in real-world videos (see Fig.~\ref{fig:qualitative_comparison}). 
These qualitative results clearly show that our method achieves significantly better performance than prior image-based end-to-end methods, further underscoring the importance of temporal cues for video-based tasks.

\textbf{Comparison with Video-based Methods.}
\begin{table}
\scriptsize
    \centering
    \setlength{\tabcolsep}{0.4mm}{
    \begin{tabular}{l|c|ccccccc|c}
    \toprule
         Method&  Bkbone&  Head&  Should.&  Elbow& Wrist&  Hip&  Knee&  Ankle& Mean\\
    \midrule
        \multicolumn{10}{c}{\textbf{Two-Stage} (Top-Down)}\\
        \midrule
        \multicolumn{10}{l}{\textit{Image-Based}}\\
         SimBase.~\cite{SimplePose_ECCV2018}&  ResNet-152&  81.7&  83.4&  80.0&  72.4&  75.3&  74.8&  67.1& 76.7\\
         HRNet~\cite{HRNet_CVPR2019}&  HRNet-W48&  82.1&  83.6&  80.4&  73.3&  75.5&  75.3&  68.5& 77.3\\
         \midrule
         \multicolumn{10}{l}{\textit{Video-Based}}\\
         PoseTrack~\cite{PoseTrack2018_CVPR2018} &  ResNet-101&  67.5&  70.2&  62.0&  51.7&  60.7&  58.7&  49.8& 60.6\\
         FastPose~\cite{fastpose_arXiv2019}&  ResNet-101&  80.0&  80.3&  69.5&  59.1&  71.4&  67.5&  59.4& 70.3\\
         STEmbed~\cite{STE_CVPR2019}&  ResNet-152&  83.8&  81.6&  77.1&  70.0&  77.4&  74.5&  70.8& 77.0\\
         PoseWarp.~\cite{PoseWarper_NIPS2019}&  HRNet-W48&  81.4&  88.3&  83.9&  78.0&  82.4&  80.5&  73.6& 81.2\\
         DCPose~\cite{DCPose_CVPR2021}&  HRNet-W48&  88.0&  88.7&  84.1&  78.4&  83.0&  81.4&  74.2& 82.8\\
         DetTrack~\cite{DetTrack_cvpr2020}&  HRNet-W48&  89.4&  89.7&  85.5&  79.5&  82.4&  80.8&  76.4& 83.8\\
         FAMIPose~\cite{FAMIPose_CVPR2022}  &  HRNet-W48&  89.1&  89.5&  84.8&  79.0&  84.2&  82.3&  74.9& 83.9\\
         TDMI~\cite{TDMI_CVPR2023}{$\dag$}  &  HRNet-W48&  90.6&  91.0&  87.2&  81.5&  85.2&  84.5&  78.7& 85.9\\
         DiffPose~\cite{DiffPose-ICCV2023}{$\dag$} &  HRNet-W48&  89.0&  91.2&  87.4&  83.5&  85.5&  87.2&  80.2& 86.4\\
         DSTA~\cite{DSTA-CVPR2024}&  ResNet-50&  87.3&  86.8&  80.0&  71.9&  78.6&  75.8&  65.4& 78.6\\
         DSTA~\cite{DSTA-CVPR2024}&  HRNet-W48&  87.6&  88.1&  84.8&  80.1&  83.6&  82.8&  75.1& 83.4\\
         DSTA~\cite{DSTA-CVPR2024}&  ViT-H&  88.1&  88.3&  86.4&  81.1&  84.2&  84.2&  76.3& 84.3\\
         \midrule
         \midrule
         \multicolumn{10}{c}{\textbf{End-to-End}} \\
         \midrule
         \multicolumn{10}{l}{\textit{Image-Based}}\\
         \rowcolor[gray]{0.9}
         PETR~\cite{PEDR-CVPR2022}&  ResNet-50& 80.5&  80.8&  71.3&  62.1&  73.4&  68.5&  61.2& 71.7\\
         \rowcolor[gray]{0.9}
         GroupPose~\cite{GroupPose-ICCV2023}&  ResNet-50&  82.4&  82.1&  73.3&  64.3&  74.4&  70.7&  63.7& 73.6\\
         \rowcolor[gray]{0.85}
         PETR~\cite{PEDR-CVPR2022}&  HRNet-W48&  82.4&  83.2&  74.4&  70.8&  74.5&  72.3& 66.9 & 75.4\\
        \rowcolor[gray]{0.85}
         GroupPose~\cite{GroupPose-ICCV2023}&  HRNet-W48&  83.3&  84.3&  77.8&  70.3&  75.6&  72.8&  66.8 & 76.3\\
         \rowcolor[gray]{0.8}
         PETR~\cite{PEDR-CVPR2022}&  Swin-L&  83.3&  84.3&  78.3&  71.3&  76.4&  73.4&  67.6 & 76.8\\
         \rowcolor[gray]{0.8}
         GroupPose~\cite{GroupPose-ICCV2023}&  Swin-L&  83.9&  84.7&  78.8&  70.6&  77.5&  74.4& 68.7 & 77.4\\
         \midrule
         \multicolumn{10}{l}{\textit{Video-Based}}\\
         \rowcolor[gray]{0.9}
         PAVE-Net (Ours)&  ResNet-50 &  86.5&  87.4&  78.9&  69.3&  78.2&  73.8&  65.8& 77.7\\
         \rowcolor[gray]{0.85}
         PAVE-Net (Ours)&  HRNet-W48&  87.1&  88.4&  80.9&  73.9&  80.3&  76.9&  69.9& 80.1\\
         \rowcolor[gray]{0.8}
         PAVE-Net (Ours)&  Swin-L& 88.2&  89.1&  81.7&  74.8&  81.6&  78.5&  71.8& {81.3} \\
        \bottomrule
    \end{tabular}
   }
    \caption{Comparison with SOTA methods on the PoseTrack2017 validation set. `$\dag$' indicates results using 4 auxiliary frames; otherwise, 2 auxiliary frames are used.}
    \label{tab:sota_compare}
    \vspace{-6mm}
\end{table}
Current state-of-the-art methods for video-based human pose estimation predominantly adopt a two-stage top-down framework: first detecting human instances frame by frame, then applying temporal modeling for single-person pose estimation. 
By explicitly focusing on individual subjects, these top-down methods have achieved superior performance. 
When using identical backbone networks for feature extraction, our proposed approach achieves results comparable to these state-of-the-art top-down methods. 
For example, with a ResNet-50 backbone, our method achieves an mAP of \textbf{77.7}, closely matching the {78.6} mAP reported by DSTA~\cite{DSTA-CVPR2024} using the same backbone. 
Our approach is flexible and readily integrates with various backbones; when using the stronger \mbox{Swin-L}~\cite{Swin-ICCV2021} backbone, our method further improves performance, reaching \textbf{81.3} mAP.

It is important to emphasize that methods leveraging temporal cues, including PoseWarper~\cite{PoseWarper_NIPS2019}, DCPose~\cite{DCPose_CVPR2021}, DetTrack~\cite{DetTrack_cvpr2020}, FAMI-Pose~\cite{FAMIPose_CVPR2022}, TDMI~\cite{TDMI_CVPR2023}, DiffPose~\cite{DiffPose-ICCV2023}, DSTA~\cite{DSTA-CVPR2024}, and our PAVE-Net, consistently outperform single-frame methods such as SimpleBaseline~\cite{SimplePose_ECCV2018}, HRNet~\cite{HRNet_CVPR2019}, PETR~\cite{PEDR-CVPR2022}, and GroupPose~\cite{GroupPose-ICCV2023}. 
This reaffirms the critical role of temporal cues from adjacent frames for more accurate, robust pose estimation in video scenarios.

\textit{Qualitative results:} Notably, existing two-stage video-based methods rely heavily on human detector performance and often struggle in challenging scenarios such as crowded environments, leading to degraded performance (see Fig.~\ref{fig:qualitative_comparison}). 
In contrast, our end-to-end video-based method removes the need for explicit human detection and consistently produces accurate and robust pose estimates even in these difficult conditions, as demonstrated in Fig.~\ref{fig:qualitative_comparison}.

\textbf{Inference Time Comparison.} 
We assess inference time, with results summarized in Table~\ref{tab:inference_time}. 
For a fair comparison, all methods use the same HRNet-W48 backbone, and two auxiliary frames are employed for all video-based approaches. 
Existing video-based MPPE methods, which follow a two-stage top-down framework, exhibit a significant increase in inference time as the number of people in the scene grows. 
In contrast, our end-to-end PAVE-Net maintains consistent inference time regardless of the number of individuals, demonstrating near-constant scalability with respect to scene complexity. 
Moreover, top-down approaches require a separate human detector, further increasing computational overhead.
As a result, compared to existing video-based methods, PAVE-Net achieves substantially lower inference time, particularly in crowded scenes. 
For example, in 20-person scenes, our method reduces inference time by \textbf{79\%} compared to DCPose~\cite{DCPose_CVPR2021} and \textbf{76\%} compared to DSTA~\cite{DSTA-CVPR2024}. 
Notably, even compared to image-based methods, our approach demonstrates comparable inference efficiency. 
Additional comparisons using other backbones (\textit{i.e.}, ResNet-50 and Swin-L) are provided in Appendix  C of the supplementary material.

This efficiency and scalability are especially valuable for industrial applications, where not only reliability but also real-time video processing is critical.

\begin{table}
\footnotesize
    \centering
    \begin{tabular}{l|ccccc}
    \toprule
         \multirow{2}{*}{Method} & \multicolumn{5}{c}{Number of Persons}\\
        &   1 &  3& 5&  10 & 20 \\
         \midrule
         \multicolumn{6}{l}{\textit{Two-Stage (Top-Down)}} \\
         DCPose~\cite{DCPose_CVPR2021} &   150&  204&  292&  431& 721\\
         DSTA~\cite{DSTA-CVPR2024}  &  122&  181&  265&  418& 631\\
    \midrule
    \multicolumn{6}{l}{\textit{End-to-End}} \\
         PETR~\cite{PEDR-CVPR2022}{$\dag$} &   \multicolumn{5}{c}{{{116}}}\\
         GroupPose~\cite{GroupPose-ICCV2023}{$\dag$} &   \multicolumn{5}{c}{{{89}}}\\
         \textbf{PAVE-Net (Ours)} &  \multicolumn{5}{c}{{{153}}}\\
    \bottomrule
    \end{tabular}
    \vspace{-2mm}
    \caption{Inference time (\textit{ms}) with HRNet-W48 backbone, measured on an A800. `{$\dag$}' denotes image-based methods.}
    \label{tab:inference_time}
\end{table}

\subsubsection{Results on the PoseTrack2018/21 Datasets\\} 
We further evaluate our model on the PoseTrack2018 and PoseTrack21 datasets. 
Due to space constraints, detailed results are provided in Appendix D of the supplementary material. 
These results clearly show that our approach consistently outperforms image-based end-to-end methods across all backbones. 
Moreover, our method achieves performance comparable to state-of-the-art two-stage video-based approaches. 
Specifically, with the ResNet-50 backbone, our method achieves \textbf{76.5} and \textbf{76.2} mAP on the two datasets, respectively. 
Using the Swin-L backbone, performance further improves by \textbf{3.6} and \textbf{3.5} points, respectively.

\subsection{Ablation Study}
We conduct ablation experiments on the PoseTrack2017 validation set to assess the impact of each component, using ResNet-50 as the backbone for all evaluations.

\begin{table}
    \centering
    \begin{tabular}{c|cc}
    \toprule
         Method & mAP & Inference Time (\textit{ms})\\
    \midrule
         Baseline & 74.5 & 336\\
         {PAVE-Net-STE} & {76.9}& {378}\\
         \textbf{PAVE-Net} & \textbf{77.7 }& \textbf{132}\\
    \bottomrule
    \end{tabular}
    \vspace{-2mm}
    \caption{Ablation of different design strategies in PAVE-Net.}
    \label{tab:ablation_baseline}
    \vspace{-5mm}
\end{table}

\textbf{Video Transformer Baseline \textit{vs.} PAVE-Net.} 
As shown in Fig.~\ref{fig:pipeline}, the video transformer baseline employs a spatiotemporal encoder (STE) to capture global dependencies among feature tokens across multiple frames. 
While straightforward, this design incurs significantly higher computational complexity and may entangle temporal features from different individuals. 
In contrast, PAVE-Net adopts a two-phase strategy: it first encodes local spatial dependencies within individual frames (SE), then employs a pose-aware spatiotemporal decoder to efficiently
aggregate features corresponding to the same individual across consecutive frames.
Table~\ref{tab:ablation_baseline} presents the performance comparison between the video transformer baseline and PAVE-Net. 
PAVE-Net achieves a substantial reduction in inference time, attaining \textbf{132} \textit{ms} versus 336 \textit{ms} for the baseline, while simultaneously improving performance by \textbf{3.2} mAP points (\textbf{77.7} vs. 74.5). Moreover, we also experimented with replacing the SE module in PAVE-Net with the STE module. This not only led to a slight drop in accuracy (\textbf{76.9}) but also substantially increased inference time to \textbf{378} \textit{ms}.
These results indicate that the SE module provides a good trade-off between accuracy and efficiency within our framework.

\textbf{Impact of Different Modules in PAVE-Net.} 
Table~\ref{tab:ablation_modules} evaluates the contribution of each module in our approach. 
PAVE-Net consists of three key components: the spatial encoder, the spatiotemporal pose decoder (STPD), and the spatiotemporal joint decoder (STJD). 
The STPD predicts 2D poses for the current frame, which are then refined by the STJD. As shown, both STPD and STJD exhibit relatively relatively few and low computational cost, making them lightweight and suitable for real-time applications.
Using only the STPD yields an mAP of 74.3, while adding the STJD further improves performance by \textbf{3.4} points, achieving 77.7 mAP. 
To further analyze the framework, we conduct an ablation where both STPD and STJD are removed. 
In this configuration, only feature tokens from the current frame, processed by the spatial encoder, are used to directly regress full-body poses. 
This reduces PAVE-Net to an image-based method where feature-to-joint misalignment tends to occur, as noted in~\cite{DisentangleReg-CVPR2021,PEDR-CVPR2022}, leading to a significant performance drop to just \textbf{61.4} mAP.

\begin{table}
    \centering
    \begin{tabular}{cc|cc|c}
    \multicolumn{2}{c|}{STPD} & \multicolumn{2}{c|}{STJD} & \multirow{3}{*}{mAP}\\
    \#Params & GFLOPs & \#Params & GFLOPs & \\
    12.34M & 7.31 & 5.43M & 9.72 & \\
    \midrule
         \multicolumn{2}{c}{\ding{55}} &  \multicolumn{2}{c|}{\ding{55}} & 61.4\\
         \multicolumn{2}{c}{\checkmark} &  \multicolumn{2}{c|}{\ding{55}} &74.3 \\
         \multicolumn{2}{c}{\checkmark} &  \multicolumn{2}{c|}{\checkmark} & \textbf{77.7}\\
    \bottomrule
    \end{tabular}
    \vspace{-2mm}
    \caption{Ablation of different modules in PAVE-Net.}
    \label{tab:ablation_modules}
\end{table}

\begin{figure}
    \centering
    \vspace{-3mm}
    \includegraphics[width=1\linewidth]{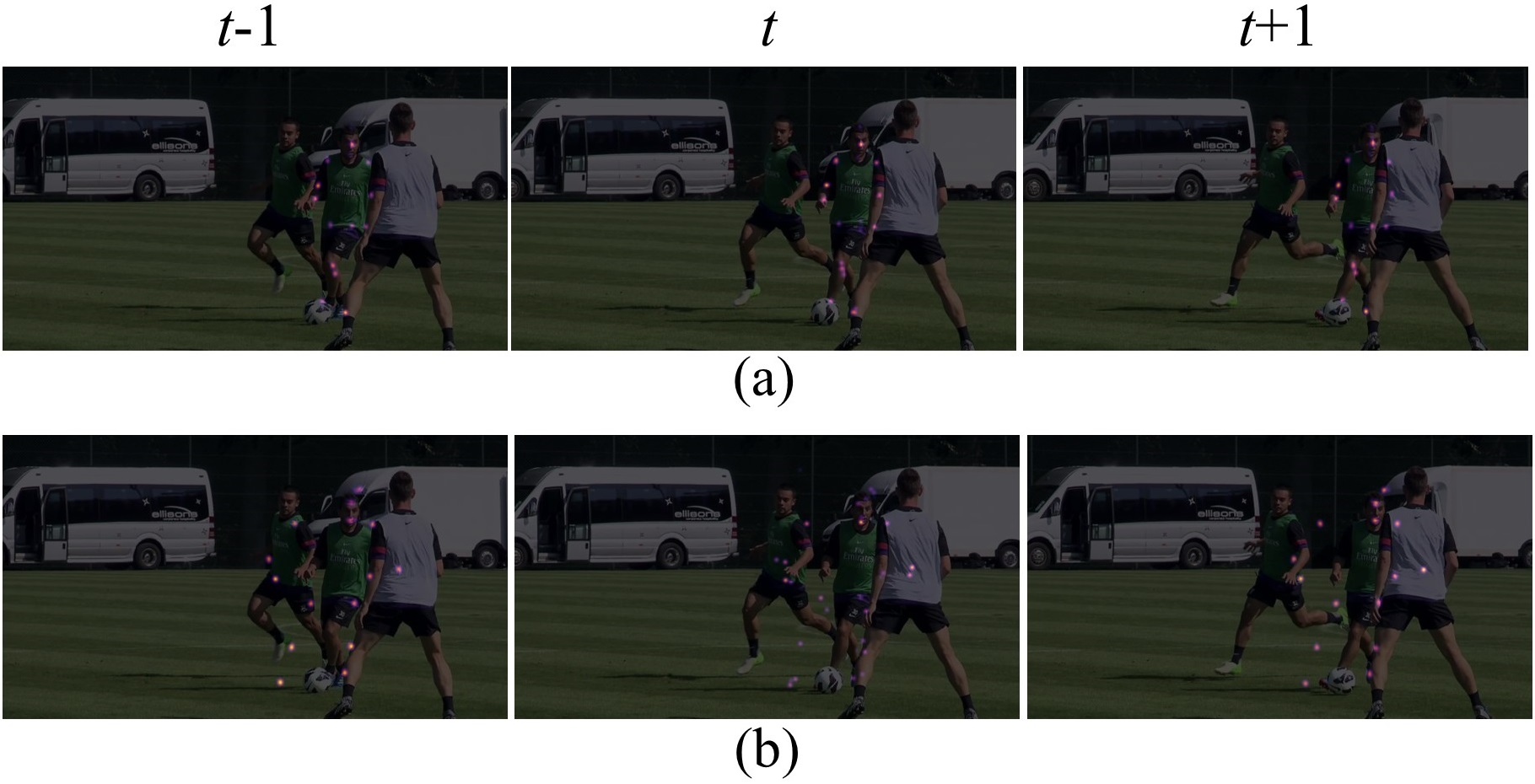}
    \vspace{-4mm}
    \caption{Features attended by the query token of the central target person across consecutive frames, highlighted with colored circles. While our pose-aware attention focuses exclusively on the target person's features (a), features from other individuals are mistakenly attended without it (b). Best viewed with zoom.
}
    \label{fig:attention_comparison}
    \vspace{-5mm}
\end{figure}

\textbf{Impact of Pose-aware Attention Mechanism.} 
In the spatiotemporal pose decoder (STPD), we introduce a novel pose-aware attention mechanism that leverages reference poses predicted from the feature tokens \( \hat{\tau}(t) \) of the current frame \( t \) to ensure that each query token attends to feature tokens associated with the same individual across multiple frames. 
We also experimented with using randomly initialized learnable parameters as reference poses, as done in~\cite{PEDR-CVPR2022}. 
As shown in Fig.~\ref{fig:attention_comparison}, pose-aware attention ensures that each query token effectively aggregates features corresponding to the same individual across consecutive frames, whereas without it, features from different individuals can become mixed. 
Consequently, as shown in Table~\ref{tab:ablation_referencePOse}, using randomly initialized learnable reference poses results in a substantial performance drop of up to \textbf{43.1} points.

These results validate that precise temporal association of the same individuals is essential for preventing feature mixing and for enabling effective temporal information aggregation. 
They also demonstrate the efficacy of our proposed pose-aware attention mechanism in establishing reliable temporal associations across frames.

\begin{table}
    \centering
    \begin{tabular}{cc}
    \toprule
         Method& mAP\\
    \midrule
         (a) & 34.6\\
         (b) & \textbf{77.7}\\
    \bottomrule
    \end{tabular}
    \caption{Different methods for generating reference poses: (a) randomly initialized learnable parameters; (b) prediction from the feature tokens \( \hat{\tau}(t) \) of the current frame.}
    \label{tab:ablation_referencePOse}
\end{table}

\begin{table}
    \centering
    \setlength{\tabcolsep}{1.5mm}{
\begin{tabular}{c|ccccc}
\toprule
Number of Layers & 1 & 2 & 3 & 4 & 5 \\
\midrule
mAP (\%)          & 67.4 & 74.8 & 77.7 & 77.2 & 77.0 \\
\bottomrule
\end{tabular}}
    \caption{Different layer numbers in STPD and STJD.}
    \label{tab:layer_number}
    \vspace{-5mm}
\end{table}

\textbf{Number of Layers in STPD and STJD.}
Table~\ref{tab:layer_number} presents the ablation study on the number of layers used in STPD and STJD. 
As the number of layers increases, model accuracy initially improves, as expected. 
However, when the number exceeds 3, the model becomes overly complex, leading to overfitting and a decline in accuracy. 
Moreover, increasing the number of layers also raises computational cost. 
Considering the trade-off between accuracy and efficiency, we choose 3 layers as the default setting.

\textbf{Auxiliary Frames.} 
In this ablation study, we examine the impact of varying the number of auxiliary frames. 
As shown in Table~\ref{tab:t_selection}, increasing auxiliary frames consistently improves performance across different backbone networks. 
This aligns with our intuition that additional frames provide complementary temporal information, thereby enhancing pose estimation accuracy for the key frame.
\begin{table}
    \centering
    \vspace{3mm}
    \begin{tabular}{c|ccc}
    \toprule
          \#Auxiliary Frame  & ResNet-50 & HRNet-W48 & Swin-L \\
    \midrule
      2 \{-1, +1\} &	77.7 & 80.1	& 81.3 \\
      4 \{-2, -1, +1, +2\} & \textbf{78.2}  & \textbf{80.5} &	\textbf{81.7} \\
    \bottomrule
    \end{tabular}
    \caption{{Different number of auxiliary frames}. `-' denotes previous frames, while `+' denotes subsequent frames.}
    \label{tab:t_selection}
    \vspace{-5mm}
\end{table}


\section{Conclusion}
We present PAVE-Net, the first fully end-to-end framework for multi-person 2D pose estimation in videos, eliminating the need for heuristic  steps such as NMS and RoI cropping. 
By combining local spatial encoding, global spatiotemporal fusion, and a pose-aware attention mechanism, PAVE-Net enables robust temporal feature aggregation across frames. 
Extensive experiments demonstrate that PAVE-Net significantly outperforms image-based end-to-end methods, while achieving accuracy comparable to state-of-the-art two-stage video-based approaches, with substantial efficiency gains.

\section{Acknowledgments}
This work was supported by ``Pioneer'' and ``Leading
Goose'' R\&D Program of Zhejiang Province (2024C01167)
and the Fundamental Research Funds for the Provincial
Universities of Zhejiang (FR24005Z).

{
    \small
    \bibliographystyle{ieeenat_fullname}
    \bibliography{main}
}

\clearpage
\setcounter{page}{1}
\maketitlesupplementary
\section*{Appendix}
In the supplementary material, we provide:\\\\
\noindent{\S}\textcolor[rgb]{1.00,0.00,0.00}{A} Additional Implementation Details. \\\\
\noindent{\S}\textcolor[rgb]{1.00,0.00,0.00}{B} Inference Time Comparison on More Backbones.\\\\
\noindent{\S}\textcolor[rgb]{1.00,0.00,0.00}{C} Experiments on PoseTrack2018/21 Datasets.\\\\
\noindent{\S}\textcolor[rgb]{1.00,0.00,0.00}{D} Qualitative Results.

\section*{A. Additional Implementation Details}
\label{sec:appendix_impl_details}

\noindent\textbf{Dataset.}
Our models are evaluated on three widely-used video-based benchmarks for human pose estimation: PoseTrack2017~\cite{PoseTrack2017_CVPR2017}, PoseTrack2018~\cite{PoseTrack2018_CVPR2018}, and PoseTrack21~\cite{PoseTrack21_CVPR2022}. These datasets collectively provide a comprehensive evaluation framework for video-based human pose estimation, covering a wide range of scenarios and challenges.
Below, we provide a detailed description of each dataset.

\begin{itemize}
    \item \textbf{PoseTrack2017}~\cite{PoseTrack2017_CVPR2017} consists of 250 video clips for training and 50 videos for validation, with a total of 80,144 pose annotations. The dataset identifies 15 keypoints per person, along with an additional label indicating joint visibility. Training videos are densely annotated in the center 30 frames, while validation videos are annotated every four frames.
    \item \textbf{PoseTrack2018}~\cite{PoseTrack2018_CVPR2018} significantly expands the scale of PoseTrack2017, containing 593 videos for training and 170 videos for validation, with a total of 153,615 pose annotations. Similar to PoseTrack2017, this dataset also identifies 15 keypoints per person and includes visibility labels. The annotation strategy remains consistent, with training videos densely annotated in the center 30 frames and validation videos annotated every four frames.
    \item \textbf{PoseTrack21}~\cite{PoseTrack21_CVPR2022} builds upon PoseTrack2018, further enriching and refining the annotations, particularly for small persons and individuals in crowded scenes. It includes a total of 177,164 human pose annotations. This dataset maintains the same keypoint structure and annotation strategy as its predecessors while addressing challenges related to occlusions and scale variations.
\end{itemize}

\noindent\textbf{Optimization.} 
Standard data augmentation techniques were applied in training, including random flipping, cropping, and scaling, with the short side constrained to the range [480, 800] and the long side not exceeding 1333. For testing, input images were resized such that the short side was fixed at 800 pixels and the long side did not exceed 1333 pixels. 
The loss weights \( \lambda_{cls} \) and \( \lambda_{rle} \) are set to 0.5 and 1.0, respectively.
We used the AdamW optimizer with a weight decay of 1\textit{e}{-4}, training for 30 epochs with a batch size of 16. The base learning rate was set to 2\textit{e}{-5}, with the backbone network's learning rate set to 2\textit{e}{-6}. The learning rate decayed by a factor of 0.1 at the 20\textit{th} and 25\textit{th} epochs. 

\begin{table}
    \centering
    \begin{tabular}{l|ccccc}
    \toprule
         \multirow{2}{*}{Method} & \multicolumn{5}{c}{Number of Persons}\\
        &   1 &  3& 5&  10 & 20 \\
         \midrule
         \multicolumn{6}{l}{\textit{Two-Stage (Top-Down)}} \\
         DCPose~\cite{DCPose_CVPR2021} &   113&  165&  216&  348& 604\\
         DSTA~\cite{DSTA-CVPR2024}  &  103&  146&  198&  337& 556\\
    \midrule
    \multicolumn{6}{l}{\textit{End-to-End}} \\
         PETR~\cite{PEDR-CVPR2022}{$\dag$} &   \multicolumn{5}{c}{{{84}}}\\
         GroupPose~\cite{GroupPose-ICCV2023}{$\dag$} &   \multicolumn{5}{c}{{{67}}}\\
         \textbf{PAVE-Net (Ours)} &  \multicolumn{5}{c}{{{132}}}\\
    \bottomrule
    \end{tabular}
    \caption{Inference time (\textit{ms}) using the ResNet-50 backbone. Measured on an A800 GPU; `{$\dag$}' indicates image-based methods.}
    \label{tab:inference_time_resnet}
\end{table}

\begin{table}
    \centering
    \begin{tabular}{l|ccccc}
    \toprule
         \multirow{2}{*}{Method} & \multicolumn{5}{c}{Number of Persons}\\
        &   1 &  3& 5&  10 & 20 \\
         \midrule
         \multicolumn{6}{l}{\textit{Two-Stage (Top-Down)}} \\
         DCPose~\cite{DCPose_CVPR2021} &   172&  231&  328&  437& 806\\
         DSTA~\cite{DSTA-CVPR2024}  &  168&  225&  294&  430& 756\\
    \midrule
    \multicolumn{6}{l}{\textit{End-to-End}} \\
         PETR~\cite{PEDR-CVPR2022}{$\dag$} &   \multicolumn{5}{c}{{{168}}}\\
         GroupPose~\cite{GroupPose-ICCV2023}{$\dag$} &   \multicolumn{5}{c}{{{146}}}\\
         \textbf{PAVE-Net (Ours)} &  \multicolumn{5}{c}{{{232}}}\\
    \bottomrule
    \end{tabular}
    \caption{Inference time (\textit{ms}) using the Swin-L backbone. Measured on an A800 GPU; `{$\dag$}' indicates image-based methods.}
    \label{tab:inference_time_swin}
    \vspace{-5mm}
\end{table}

\section*{B. Inference Time Comparison on More Backbones.}
\label{sec:appendix_computation complexity}
Tables~\ref{tab:inference_time_resnet} and ~\ref{tab:inference_time_swin} present the inference time results of various methods using the ResNet-50 and Swin-L backbones. Despite leveraging multi-frame information, our approach achieves comparable efficiency in inference time to image-based end-to-end methods. Furthermore, compared to existing video-based methods, which rely on a two-stage top-down framework, our PAVE-Net significantly reduces inference time, particularly in multi-person scenarios. For instance, in the 20-person setting, our method reduces inference time by \textbf{79\%} (or \textbf{72\%}) compared to DCPose~\cite{DCPose_CVPR2021} when using the ResNet-50 (or Swin-L) backbone. This efficiency is highly valuable for industrial applications, where real-time video processing is crucial.

\begin{table}
    \scriptsize
        \centering
        \setlength{\tabcolsep}{0.4mm}{
        \begin{tabular}{l|c|ccccccc|c}
        \toprule
        \toprule
        Method&  Bkbone&  Head&  Should.&  Elbow& Wrist&  Hip&  Knee&  Ankle& Mean\\
        \midrule
        \multicolumn{10}{c}{\textbf{Two-Stage} (Top-Down)}\\
            \midrule
            \multicolumn{10}{l}{\textit{Image-Based}}\\
             AlphaPose~\cite{AlphaPose}&  ResNet-50&  63.9&  78.7&  77.4&  71.0&  73.7&  73.0&  69.7& 71.9\\
             MDPN~\cite{MDPN_ECCV-W2018}&  ResNet-152&  75.4&  81.2&  79.0&  74.1&  72.4&  73.0&  69.9& 75.0\\
             \midrule
             \multicolumn{10}{l}{\textit{Video-Based}}\\
             PoseWarp.~\cite{PoseWarper_NIPS2019}&  HRNet-W48&  79.9&  86.3&  82.4&  77.5&  79.8&  78.8&  73.2& 79.7\\
             DCPose~\cite{DCPose_CVPR2021}&  HRNet-W48&  84.0&  88.8&  86.2&  79.4&  72.0&  80.6&  76.2& 80.9\\
             DetTrack~\cite{DetTrack_cvpr2020}&  HRNet-W48&  84.9&  87.4&  84.8&  79.2&  77.6&  79.7&  75.3& 81.5\\
             FAMIPose~\cite{FAMIPose_CVPR2022}{$\dag$}&  HRNet-W48&  85.5& 87.7&  84.2&  79.2&  81.4&  81.1&  74.9& 82.2\\
             TDMI~\cite{TDMI_CVPR2023}{$\dag$}&  HRNet-W48&  86.7&  88.9&  85.4&  80.6&  82.4&  82.1&  77.6& 83.6\\
             DiffPose~\cite{DiffPose-ICCV2023}{$\dag$} &  HRNet-W48&  85.0&  87.7&  84.3&  81.5&  81.4&  82.9&  77.6& 83.0\\
             DSTA~\cite{DSTA-CVPR2024}&  ResNet-152&  84.8&  86.7&  80.2&  72.1&  78.5&  77.2&  67.3& 78.5\\
             DSTA~\cite{DSTA-CVPR2024}&  HRNet-W48&  85.7&  88.1&  82.9&  76.2&  81.2&  78.0&  72.2& 80.9\\
             DSTA~\cite{DSTA-CVPR2024}&  ViT-H&  85.1&  87.6&  84.5&  80.1&  79.8&  81.1&  75.4& 82.1\\
             \midrule
             \midrule
             \multicolumn{10}{c}{\textbf{End-to-End}} \\
             \midrule
             \multicolumn{10}{l}{\textit{Image-Based}}\\
             \rowcolor[gray]{0.9}
             PETR~\cite{PEDR-CVPR2022}&  ResNet-50&  80.1&  79.5&  69.8&  61.6&  72.3&  66.1&  59.6& 70.5\\
             \rowcolor[gray]{0.9}
             GroupPose~\cite{GroupPose-ICCV2023}&  ResNet-50&  81.2&  81.5&  71.8&  62.8&  72.3&  68.6&  62.1& 72.1\\
             \rowcolor[gray]{0.85}
             PETR~\cite{PEDR-CVPR2022}&  HRNet-W48&  81.6&  81.7&  74.3&  70.9&  72.3&  70.2&  64.3& 74.1\\
             \rowcolor[gray]{0.85}
             GroupPose~\cite{GroupPose-ICCV2023}&  HRNet-W48&  82.1&  83.8&  77.2&  69.8&  74.1&  71.9&  65.6& 75.4\\
             \rowcolor[gray]{0.8}
             PETR~\cite{PEDR-CVPR2022}&  Swin-L&     82.2&  83.2&  76.4&  70.6&  74.8&  71.9&  66.1& 75.5\\
             \rowcolor[gray]{0.8}
             GroupPose~\cite{GroupPose-ICCV2023}&  Swin-L&     82.3&  83.4&  78.1&  70.7&  74.8&  72.4&  68.2& 76.1\\
             \midrule
             \multicolumn{10}{l}{\textit{Video-Based}}\\
             \rowcolor[gray]{0.9}
             PAVE-Net (Ours)&  ResNet-50 &  83.2&  84.7&  77.8&  69.9&  74.1&  72.7&  66.8&  76.1\\
             \rowcolor[gray]{0.85}
             PAVE-Net (Ours)&  HRNet-W48&   83.8&  85.6&  81.2&  74.3&  77.6&  74.7&  70.5&  78.6\\
             \rowcolor[gray]{0.8}
             PAVE-Net (Ours)&  Swin-L&      84.9&  87.5&  81.7&  74.5&  77.9&  76.9&  72.4& \textbf{79.7} \\
            \bottomrule
            \bottomrule
        \end{tabular}
       }
    \caption{Comparison with the SOTAs on PoseTrack2018 val.set. `$\dag$' indicates results obtained using 4 auxiliary frames; otherwise, only 2 auxiliary frames are employed.}
        \label{tab:sota_compare_posetrack2018}
\end{table}

\section*{C. Experiments on PoseTrack2018/21 Datasets}
Tables~\ref{tab:sota_compare_posetrack2018} and \ref{tab:sota_compare_posetrack21} present the comparisons of our method with the state-of-the-art methods on the PoseTrack2018 and PoseTrack21 validation sets, respectively.
These results further confirm that our approach consistently outperforms image-based end-to-end methods across all backbone networks while achieving performance comparable to state-of-the-art two-stage video-based approaches.



\begin{table}
  \scriptsize
      \centering
      \setlength{\tabcolsep}{0.4mm}{
      \begin{tabular}{l|c|ccccccc|c}
      \toprule
      \toprule
      Method&  Bkbone&  Head&  Should.&  Elbow& Wrist&  Hip&  Knee&  Ankle& Mean\\
      \midrule
         \multicolumn{10}{c}{\textbf{Two-Stage} (Top-Down)}\\
          \midrule
          \multicolumn{10}{l}{\textit{Image-Based}}\\
           SimBase.~\cite{SimplePose_ECCV2018}&  ResNet-152&  80.5&  81.2&  73.2&  64.8&  73.9&  72.7&  67.7& 73.9\\
           HRNet~\cite{HRNet_CVPR2019}&  HRNet-W48&  81.5&  83.2&  81.1&  75.4&  79.2&  77.8&  71.9& 78.8\\
           \midrule
           \multicolumn{10}{l}{\textit{Video-Based}}\\
           PoseWarp.~\cite{PoseWarper_NIPS2019}&  HRNet-W48&  82.3&  84.0&  82.2&  75.5&  80.7&  78.7&  71.6& 79.5\\
           DCPose~\cite{DCPose_CVPR2021}&         HRNet-W48&  83.7&  84.4&  82.6&  78.7&  80.1&  79.8&  74.4& 80.7\\
           FAMIPose~\cite{FAMIPose_CVPR2022}{$\dag$}&     HRNet-W48&  83.3&  85.4&  82.9&  78.6&  81.3&  80.5&  75.3& 81.2\\
           TDMI~\cite{TDMI_CVPR2023}{$\dag$}&  HRNet-W48&  86.8&  87.4&  85.1&  81.4&  83.8&  82.7&  78.0& 83.8\\
           DiffPose~\cite{DiffPose-ICCV2023}{$\dag$} &  HRNet-W48&  84.7&  85.6&  83.6&  80.8&  81.4&  83.5&  80.0& 82.9\\
           DSTA~\cite{DSTA-CVPR2024}&  ResNet-152&  85.7&  84.6&  79.8&  73.2&  78.5&  75.1&  68.6& 78.4\\
           DSTA~\cite{DSTA-CVPR2024}&  HRNet-W48&  85.3&  84.8&  81.9&  78.6&  81.6&  77.8&  73.4& 80.8\\
           DSTA~\cite{DSTA-CVPR2024}&  ViT-H&  86.4&  85.3&  82.1&  80.6&  81.2&  81.5&  76.2& 82.2\\
           \midrule
           \midrule
           \multicolumn{10}{c}{\textbf{End-to-End}} \\
           \midrule
           \multicolumn{10}{l}{\textit{Image-Based}}\\
           \rowcolor[gray]{0.9}
           PETR~\cite{PEDR-CVPR2022}&  ResNet-50&  80.3&  79.1&  69.4&  61.2&  71.7&  65.9&  59.8& 70.3\\
           \rowcolor[gray]{0.9}
           GroupPose~\cite{GroupPose-ICCV2023}&  ResNet-50&  80.8 & 80.3 & 71.2 & 63.2 & 73.1 & 68.8 & 61.8 & 71.9 \\
           \rowcolor[gray]{0.85}
           PETR~\cite{PEDR-CVPR2022}&  HRNet-W48&  81.1&  80.8&  74.8&  70.0&  71.3&  70.5&  63.3& 73.6\\
           \rowcolor[gray]{0.85}
           GroupPose~\cite{GroupPose-ICCV2023}&  HRNet-W48&  82.7 & 84.6 & 76.6 & 69.2 & 74.0 & 71.2 & 65.4 & 75.3 \\
           \rowcolor[gray]{0.8}
           PETR~\cite{PEDR-CVPR2022}&  Swin-L&     81.6&  83.3&  75.4&  70.7&  74.3&  71.2&  66.0& 75.1\\
           \rowcolor[gray]{0.8}
           GroupPose~\cite{GroupPose-ICCV2023}&  Swin-L&     82.4 & 82.9 & 77.6 & 70.3& 73.7 & 72.5 & 67.4 & 75.7 \\
           \midrule
           \multicolumn{10}{l}{\textit{Video-Based}}\\
           \rowcolor[gray]{0.9}
           PAVE-Net (Ours)&  ResNet-50 &  82.6 & 82.8 & 77.2 & 67.6 & 76.8 & 72.8 & 68.2 & 75.9 \\
           \rowcolor[gray]{0.85}
           PAVE-Net (Ours)&  HRNet-W48&   83.2 & 84.5 & 81.6 & 74.5 & 78.1 & 74.6 & 71.5 & 78.6 \\
           \rowcolor[gray]{0.8}
           PAVE-Net (Ours)&  Swin-L&      84.7 & 86.5 & 81.9 & 74.7 & 77.4 & 76.6 & 71.4 & \textbf{79.4} \\
          \bottomrule
          \bottomrule
      \end{tabular}
     }
     \caption{Comparison with the SOTAs on PoseTrack21 val.set. `$\dag$' indicates results obtained using 4 auxiliary frames; otherwise, only 2 auxiliary frames are employed.}
      \label{tab:sota_compare_posetrack21}
  \end{table}
  

\begin{figure*}
    \centering
    \includegraphics[width=1.0\linewidth]{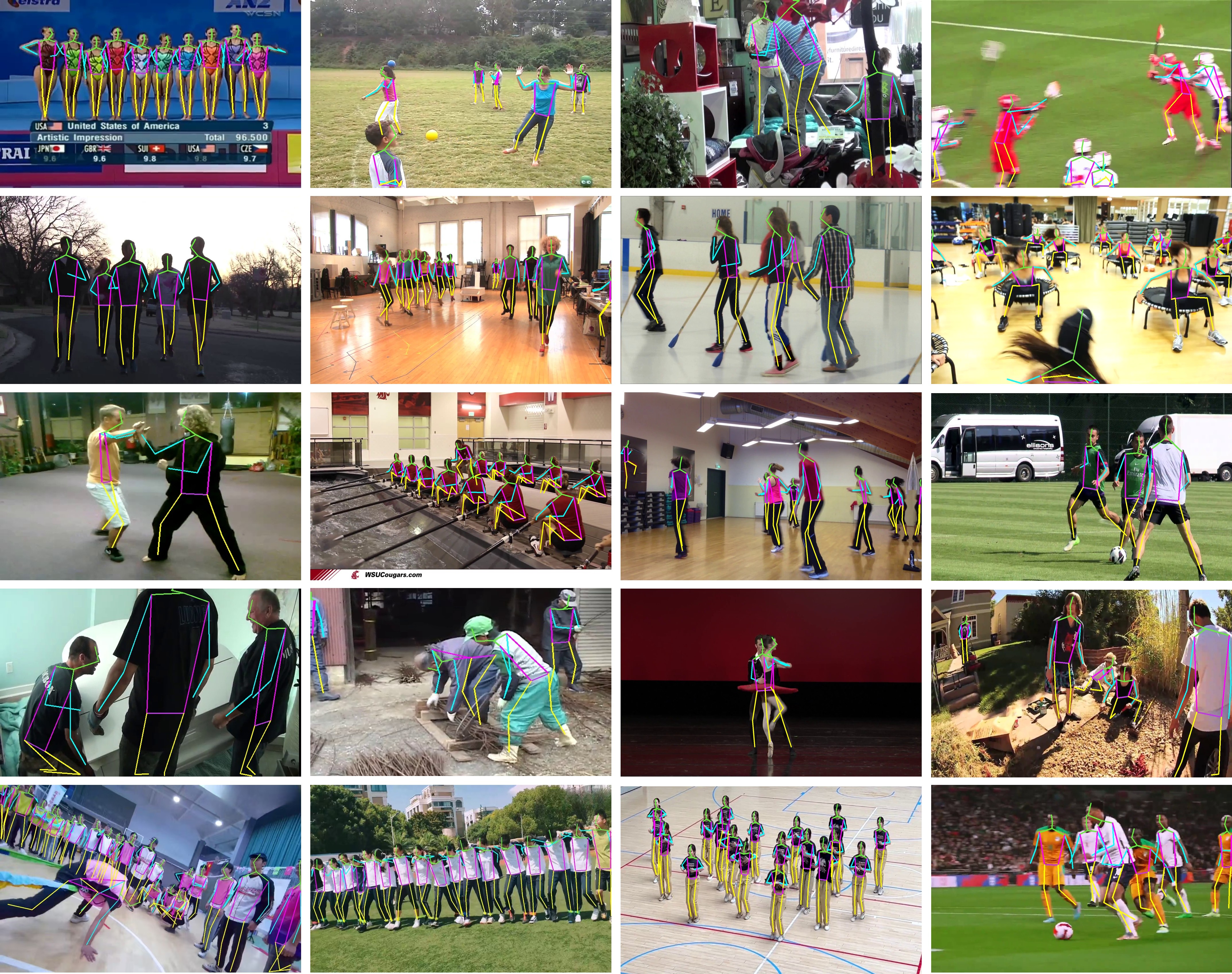}
    \caption{{Additional qualitative results} of our method on the PoseTrack validation sets and in-the-wild videos.}
    \label{fig:qualitative_result_supplementary}
\end{figure*}

\section*{D. Qualitative Results}
Additional qualitative results on the PoseTrack validation sets and in-the-wild videos are presented in Fig.~\ref{fig:qualitative_result_supplementary}. Further results are available in the accompanying video material.

\end{document}